\documentclass[letterpaper, 10 pt, journal, twoside]{IEEEtran}

\usepackage{graphics}           
\usepackage{times}              
\usepackage{amsmath}            
\usepackage{amssymb}            
\usepackage{graphicx}
\usepackage{algorithm}
\usepackage[noend]{algpseudocode}
\usepackage{booktabs}
\usepackage{color}
\usepackage[nocompress]{cite} 
\usepackage{multirow}
\usepackage{subfig}
\usepackage{hhline}
\usepackage{hyperref}
\usepackage{xcolor}

\usepackage{siunitx}
\usepackage{scalerel}
\usepackage{svg}
\usepackage{soul}

\definecolor{instructioncolor}{rgb}{.5,.5,.5}

\usepackage[font=small]{caption}

\def\eqref#1{Eq.~(\ref{#1})}

\makeatletter
\usepackage{xspace}
\DeclareRobustCommand\onedot{\futurelet\@let@token\@onedot}
\def\@onedot{\ifx\@let@token.\else.\null\fi\xspace}

\makeatother

\usepackage{array}
\newcolumntype{L}[1]{>{\raggedright\let\newline\\\arraybackslash\hspace{0pt}}m{#1}}
\newcolumntype{C}[1]{>{\centering\let\newline\\\arraybackslash\hspace{0pt}}m{#1}}
\newcolumntype{R}[1]{>{\raggedleft\let\newline\\\arraybackslash\hspace{0pt}}m{#1}}

\newcommand{\degrees}{{\mbox{$^\circ$}}}

\usepackage[normalem]{ulem}
\usepackage{pifont}          
\usepackage{comment}
\usepackage{flushend}
\newif\ifdoubleblind
\doubleblindfalse 
\title{Dynamic-ICP: Doppler-Aware Iterative Closest Point Registration for Dynamic Scenes}
\ifdoubleblind
  \author{}
\else
\author{Dong Wang, Daniel Casado Herraez, Stefan May and Andreas Nüchter 
\thanks{Manuscript received: October 21, 2025; Revised: January 26, 2026; Accepted: February 23, 2026. This paper was recommended for publication by Editor Javier Civera upon evaluation of the Associate Editor and Reviewers' comments. This work was in parts supported by the Federal Ministry for Economic Affairs and Climate Action (BMWK) on the basis of a decision by the German Bundestag under the grant number KK5150106RL4.}%
\thanks{D. Wang {\tt\small (dong.wang@uni-wuerzburg.de)} and A. Nüchter are with Julius-Maximilians-Universität Würzburg, Germany. Andreas Nüchter is also with the Zentrum für Telematik e.V., Würzburg and currently International Visiting Chair at U2IS, ENSTA, Institut Polytechnique de Paris, France. S. May is with Nuremberg Institute of Technology Georg Simon Ohm, Germany. D. Casado Herraez is with CARIAD SE and the University of Bonn, Germany.} 
\thanks{Digital Object Identifier (DOI): see top of this page.}}
\fi
\begin{document}
\maketitle
\markboth{IEEE Robotics and Automation Letters. Preprint Version. Accepted February, 2026}
{Wang \MakeLowercase{\textit{et al.}}: Dynamic-ICP}

\begin{abstract}
Reliable odometry in highly dynamic environments remains challenging when it relies on ICP-based registration: ICP assumes near-static scenes and degrades in repetitive or low-texture geometry. We introduce Dynamic-ICP, a Doppler-aware registration framework. The method (i) estimates ego translational velocity from per-point Doppler velocity via robust regression and builds a velocity filter, (ii) clusters dynamic objects and reconstructs object-wise translational velocities from ego-compensated radial measurements, (iii) predicts dynamic points with a constant-velocity model, and (iv) aligns scans using a compact objective that combines point-to-plane geometry residual with a translation-invariant, rotation-only Doppler residual. The approach requires no external sensors or sensor–vehicle calibration and operates directly on FMCW LiDAR range and Doppler velocities. We evaluate Dynamic-ICP on three real-world datasets-HeRCULES, HeLiPR, AevaScenes-focusing on highly dynamic scenes. Dynamic-ICP consistently improves rotational stability and translation accuracy over the state-of-the-art methods. To encourage further research, the code is available at: 
\ifdoubleblind
\url{https://anonymous.4open.science/r/Dynamic-ICP-62B5}. 
\else
\url{https://github.com/JMUWRobotics/Dynamic-ICP}. 
\fi
\end{abstract}
\begin{IEEEkeywords}
  Odometry, Mapping, Localization, SLAM, Autonomous Vehicle Navigation
\end{IEEEkeywords}

\section{INTRODUCTION}
\label{sec:intro}
\IEEEPARstart{R}{eliable} odometry in unknown environments is a fundamental requirement for robust autonomy in ground, aerial, and mobile robots. Decades of research have yielded numerous point cloud registration methods for odometry pipelines, among which the Iterative Closest Point (ICP)~\cite{besl1992method}~\cite{chen1992object} has been widely adopted due to its precision and efficiency. ICP aligns a source to a target cloud by alternating correspondence search with transform refinement to minimize the geometric error. While ICP variants can be highly accurate in quasistatic scenes, they implicitly assume that most points are stationary between successive frames, an assumption violated in highly dynamic settings. As a result, moving objects induce spurious correspondences and bias, and performance further degrades in geometrically repetitive, low-texture environments (e.g., tunnels, bridges) where ambiguous matches can lead to drift.

In recent years, frequency-modulated continuous-wave (FMCW) LiDAR technology has advanced beyond traditional LiDAR systems by providing range and per-point Doppler velocity. These sensors measure direct motion cues in dynamic scenes by comparing phase shifts across the transmitted chirp sequence. However, the use of Doppler for understanding dynamic scenes is still in early stages.

\begin{figure}[t]
  \centering
  \includegraphics[width=\linewidth]{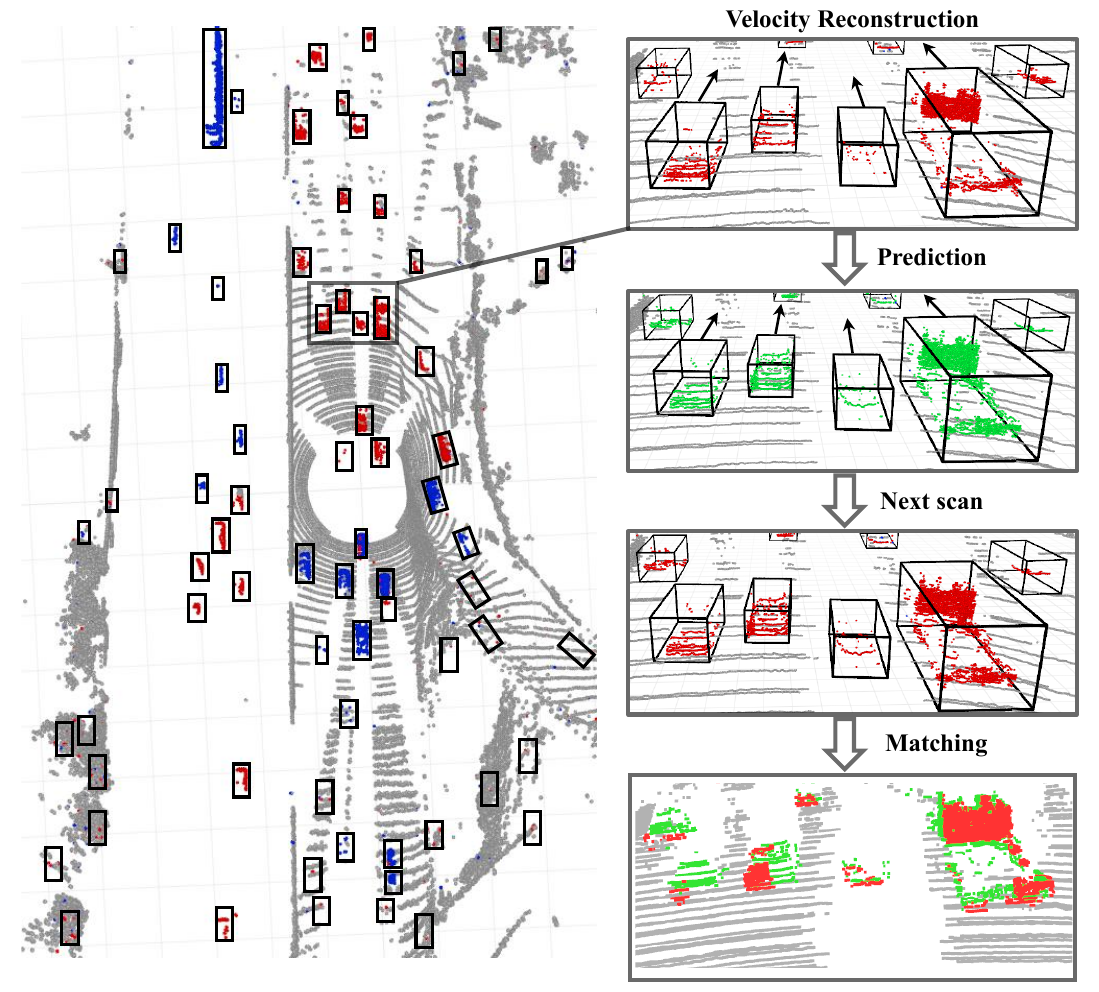}
  \vspace{-0.2cm} 
  \caption{Workflow of Dynamic-ICP for dynamic objects. Left: FMCW LiDAR scan in highway scenarios. Dynamic points are colored and clustered. Right: Velocity reconstruction, prediction, and matching of dynamic objects. The raw dynamic points are colored \textcolor{red}{red}, while the predicted points are colored \textcolor{green}{green}. Black arrows and boxes represent the object's velocity and bounding box, respectively.}
  \label{fig:motivation}
  \vspace{0.1em}
\end{figure}

To address these limitations, we propose Dynamic-ICP, a Doppler-aware variant of ICP that explicitly models scene dynamics and fast ego motion. Rather than assuming a fully static scene, Dynamic-ICP reduces sensitivity to dynamics by predicting where points will be in the next frame before establishing correspondences. Concretely, we exploit per-point Doppler velocity measurements (e.g., from FMCW sensors) to cluster moving objects, reconstruct each object's full 3D velocity, and warp points into their next-frame states, as shown in Fig.~\ref {fig:motivation}. A distance-adaptive dynamic correlation then guides correspondence weighting, suppressing spurious matches from dynamic objects while preserving informative structure.
Compared to approaches that simply reject dynamic points or rely solely on robust loss functions~\cite{behley2018efficient}~\cite{milioto2019rangenet++}~\cite{zhang2014loam}, Dynamic-ICP prioritizes Doppler-consistent pairs and performs correspondence search in a motion-compensated domain. The added rotation-only Doppler residual eliminates classic ICP degeneracies and reduces drift in repetitive or low-texture regions. In highly dynamic scenes, per-cluster velocity prediction aligns moving objects across frames. It also stabilizes orientation under strong ego and object motion. All of this is done while retaining the simplicity and real-time efficiency that makes ICP practical.

We evaluate Dynamic-ICP across diverse, highly dynamic scenarios with substantial object motion. Experiments demonstrate consistent gains in rotation/translation accuracy, faster convergence, and improved robustness under severe dynamics. We also report detailed ablations to quantify the effect of Doppler-based clustering, velocity reconstruction, and distance-adaptive correlation on overall performance. Code will be released to facilitate reproduction and extension. To summarize, our contributions are:
\begin{itemize}
    \item We present Dynamic-ICP, a Doppler-aware dynamic ICP designed for highly dynamic scenes where standard ICP degrades.
    \item We leverage per-point Doppler velocity to cluster moving objects and reconstruct each object's translational velocity. A distance-adaptive motion correlation then extrapolates object states to the next frame to guide correspondences.
    \item Our Doppler-consistent matching substantially alleviates the rotation-estimation failure modes of vanilla ICP, improving stability and accuracy under strong ego- and object motion.
    \item Extensive experiments across diverse datasets show state-of-the-art performance in highly dynamic scenarios. 
\end{itemize}



\section{RELATED WORK}
\label{sec:related}
In this section, we review state-of-the-art ICP approaches, including traditional frameworks and recent methods that leverage Doppler information. We also discuss techniques that integrate ICP with scene flow estimation and present our method in the context of Doppler-aware ICP for dynamic scenes. 
\subsection{Iterative Closest Point and Variants}
Iterative Closest Point (ICP) is the canonical method for rigid point cloud alignment, originating from early formulations that alternated between closest-point association and least-squares pose estimation~\cite{besl1992method}~\cite{chen1992object}. Efficiency and robustness have been improved through careful choices of metrics, sampling, and weighting~\cite{nuchter2007cached}~\cite{rusinkiewicz2001efficient}, probabilistic modeling as in Generalized-ICP~\cite{segal2009generalized}, and outlier handling via trimmed objectives~\cite{chetverikov2002trimmed}. Distribution-based registration such as the normal distributions transform, replaces discrete matches with continuous densities to enhance convergence in sparse or noisy settings~\cite{biber2003normal}~\cite{magnusson2009three}. To extend ICP to large baselines, global or certifiable methods provide strong initializations and outlier guarantees~\cite{yang2020teaser}~\cite{zhou2016fast}. Recent variants target real-time and degenerate scenes: Voxelized GICP accelerates and stabilizes GICP with voxel-wise covariance aggregation~\cite{koide2021voxelized}, and KISS-ICP~\cite{vizzo2023kiss} shows that a carefully engineered point-to-point pipeline with simple motion compensation and robust thresholds can be both accurate and broadly applicable across sensors. Despite these advances, most ICP-style pipelines implicitly assume near-static scenes over short horizons, which limits performance under strong ego motion and highly dynamic environments.
\subsection{Doppler Velocity based Matching}
FMCW LiDAR sensors provide per-point radial velocity via Doppler shift, offering direct motion cues that can disambiguate correspondences under strong ego and object motion. Previous work has exploited Doppler velocity for instantaneous estimation of ego-motion, using the radial velocity constraint to separate static structures from moving objects and to infer velocity vectors~\cite{kellner2013instantaneous}~\cite{wang2025doppler}~\cite{zhao2024fmcw}~\cite{zhao2024free}. Incorporating Doppler velocity into registration has been explored by constraining or guiding correspondences based on velocity consistency. This approach is sometimes referred to as Doppler-constrained ICP or motion-compensated matching~\cite{herraez2024radar}~\cite{herraez2025rai}~\cite{hexsel2022dicp}~\cite{wu2022picking}. In the SLAM framework, the Doppler velocity sensor is integrated with inertial and other exteroceptive sensors to stabilize odometry, reduce drift, and improve data association in challenging conditions, such as rain, low texture, or repeated geometry~\cite{yoon2023need}. However, most existing methods rely on predominantly static environments and treat moving points as outliers. In contrast, our Doppler-aware matching method reconstructs 3D velocities for each object from the Doppler effect and predicts their positions in the next frame. This method yields more reliable correspondences and stronger rotational constraints in highly dynamic scenes.
\subsection{Scene Flow}
Scene flow estimates a dense 3D motion field between consecutive observations, generalizing optical flow to three dimensions~\cite{vedula1999three}. Classical formulations solved for per-point 3D motion under smoothness constraints, while modern learning-based methods infer flow directly from point clouds using cost volumes, permutation-invariant layers, and optimal transport objectives~\cite{gu2019hplflownet}~\cite{kittenplon2021flowstep3d}~\cite{liu2019flownet3d}. Recent directions relevant to dynamic registration include radar-based cross-modal supervision for 4D radar scene flow~\cite{ding2023hidden}, ICP-Flow~\cite{lin2024icp}, which uses rigid-motion priors and classical ICP to produce consistent object-wise flow without training, and 4D voxel networks that fuse multiple frames for efficient spatio-temporal reasoning~\cite{kim2025flow4d}. Complementary to these, cross-modal Doppler guidance has been proposed to transfer radar-derived 3D velocities to LiDAR for self-supervised scene flow~\cite{khoche2025dogflow}. However, dense flow estimation can be computationally demanding and may rely on learned priors that do not transfer across sensors or environments. In contrast, Dynamic-ICP provides a lightweight, training-free alternative to dense scene flow that retains real-time efficiency in highly dynamic scenes.

Inspired by previous works Doppler-ICP~\cite{hexsel2022dicp}, DoGFlow~\cite{khoche2025dogflow}, Flow4d~\cite{kim2025flow4d}, ICP-Flow~\cite{lin2024icp} and Doppler-SLAM~\cite{wang2025doppler}, we propose Dynamic-ICP, a Doppler-aware registration framework that integrates per-point Doppler velocity into correspondence search. Rather than masking dynamic content or relying on dense scene flow estimation, Dynamic-ICP reconstructs object-wise 3D velocities from Doppler velocity, predicts next-frame positions with a distance-adaptive correlation model, and performs ICP in this motion-compensated space. This design yields more reliable correspondences and improved rotation and translation estimates in high-speed, highly dynamic scenes while preserving the simplicity and efficiency of ICP. Another advantage of our method is that it operates without requiring extrinsic sensor-vehicle calibration compared to Doppler-ICP~\cite{hexsel2022dicp}. In summary, Dynamic-ICP combines Doppler-aware matching and ICP-style optimization to provide lightweight, robust registration for classic ICP's failure modes in dynamic scenes.
\section{Dynamic-ICP}
    \begin{figure*}[ht]
        \centering
        \includegraphics[width=1\linewidth]{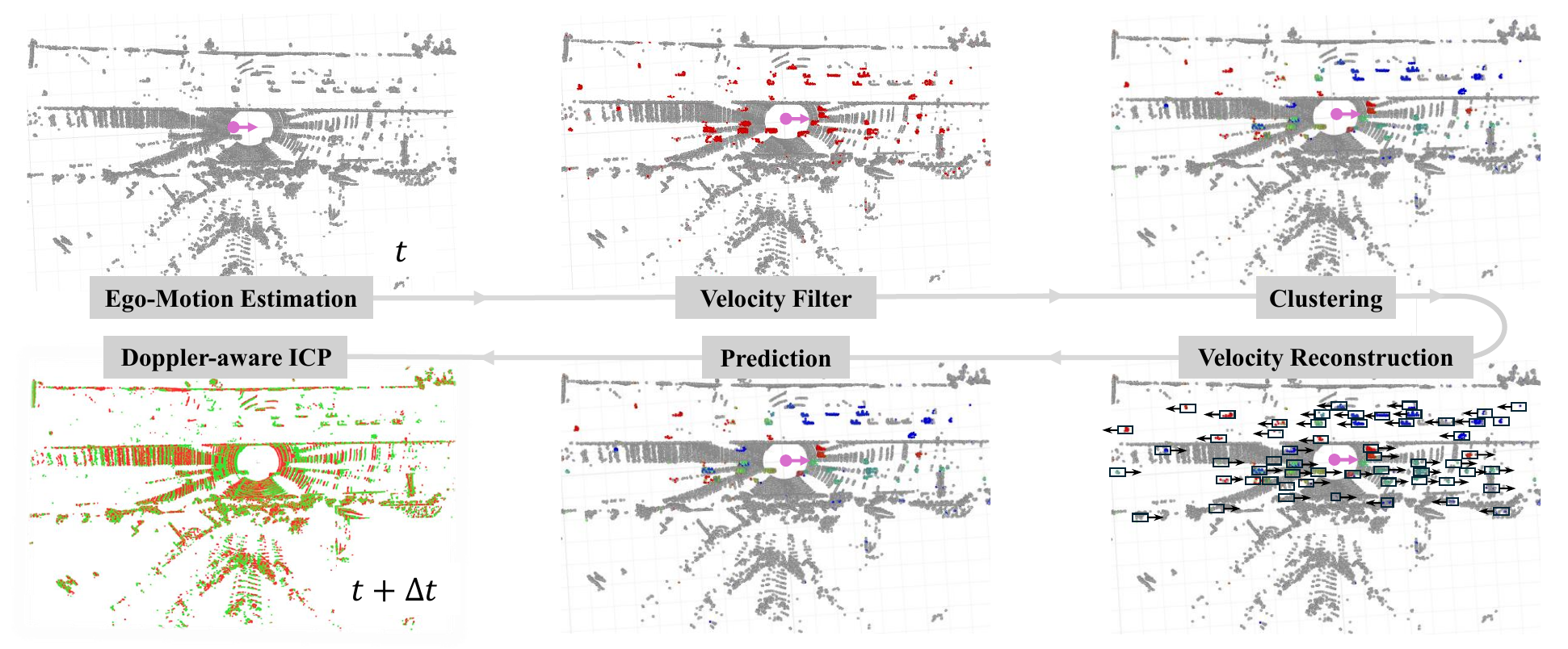}
        \caption{\small{Pipeline of Dynamic-ICP consists of four main modules: (i) Ego-Motion Estimation (Sec.~\ref{sec:ego}); (ii) Dynamic Points Clustering (Velocity Filter, Clustering and Velocity Reconstruction) (Sec.~\ref{sec:cluster}); (iii) Dynamic Points Prediction (Sec.~\ref{sec:predict}); and (iv) Doppler-aware ICP Matching (Sec.~\ref{sec:match}). The figure illustrates the workflow on point cloud data: starting from a raw scan, ego velocity (arrows) is estimated from per-point Doppler velocity. The velocity filter distinguishes dynamic points (\textcolor{red}{red}) from the static background (\textcolor{gray}{gray}) and clusters the dynamic set into individual objects (colored). For each cluster, the object velocity is reconstructed from its points’ Doppler velocities, yielding a velocity vector and a bounding box (black arrow and box). These velocities are then used to predict object states to the next frame. Finally, the scan (\textcolor{green}{green}) at time~$t$ (predicted dynamic points, together with the static points) are aligned to the scan (\textcolor{red}{red}) at time~$t + \Delta t$ via Doppler-aware ICP.}}
        \label{fig:overview}
    \end{figure*}
We use standard notation: vectors are bold lowercase (e.g.,~$\boldsymbol{b}$), matrices bold uppercase (e.g.,~$\boldsymbol{B}$), and frames by superscripts/subscripts (e.g.,~$\boldsymbol{b}^l$). The rigid transform is~$\boldsymbol{T}=[\boldsymbol{R}\,|\,\boldsymbol{t}]$ with~$\boldsymbol{R}\in SO(3)$ and~$\boldsymbol{t}\in\mathbb{R}^3$.
\subsection{Problem Statement}
We denote the sensor frame for the previous point cloud as the source frame,~$\mathcal{F}_S$, and the current point cloud as the target frame,~$\mathcal{F}_T$. Let 
\begin{equation}
\boldsymbol{u}_i^t \;\triangleq\; \frac{\boldsymbol{p}_i^t}{\|\boldsymbol{p}_i^t\|},  \mathcal{P}_t=\{( \boldsymbol{p}_i^t,\;\boldsymbol{u}_i^t,\; s_i^t )\}_{i=1}^{N_t}, 
  \label{eq:eq1}
\end{equation}
be the point set acquired at time~$t$, where ~$\boldsymbol{p}_i^t\in\mathbb{R}^3$ is the position of the point in the sensor frame,~\(\boldsymbol{u}_i^t\in\mathbb{S}^2\) is the unit line-of-sight (LOS) vector from the sensor to \(\boldsymbol{p}_i^t\), and \(s_i^t\in\mathbb{R}\) is the measured Doppler velocity along \(\boldsymbol{u}_i^t\) (positive away from the sensor). Likewise \(\mathcal{P}_{t+1}\) is available at time ~$t{+}1$.
Our goals are to:
\begin{enumerate}
    \item estimate the transformation \( \boldsymbol{T}_{t\rightarrow t+1}\in SE(3) \) that aligns \(\mathcal{P}_t\) to \(\mathcal{P}_{t+1}\);
    \item achieve registration that remains reliable under strong ego- and object motion.
\end{enumerate}
We write \( \boldsymbol{T}=[\boldsymbol{R}\;|\;\boldsymbol{t}] \) with \(\boldsymbol{R}\in SO(3)\) and \(\boldsymbol{t}\in\mathbb{R}^3\). A body twist is \(\boldsymbol{\xi}=[\boldsymbol{\omega}^\top\;\boldsymbol{v}^\top]^\top\in\mathbb{R}^6\).
\subsection{System Overview}
Fig.~\ref{fig:overview} illustrates the overall system architecture, highlighting four key modules: (i) Ego-Motion Estimation, which estimates the ego velocity from Doppler velocity, (ii) Dynamic Points Clustering, which clusters the dynamic points into individual objects, (iii) Dynamic Points Prediction, predicting the dynamic points in next frame, and (iv) Doppler-aware ICP Matching. The following subsections describe the design and implementation of each module in detail.
\subsection{Ego-Motion Estimation}\label{sec:ego}
To estimate the ego-motion of the sensor, we first leverage the static points. For static scene points, the measured Doppler velocity equals the velocity induced by the ego-motion at the point projected onto the LOS:
\begin{equation}
s_i^t = -\,(\boldsymbol{u}_i^t)^\top \big(\boldsymbol{v} + \boldsymbol{\omega}\times \boldsymbol{p}_i^t\big),
\label{eq:radial_static}
\end{equation}
where~\(\boldsymbol{v},\boldsymbol{\omega}\) are the instantaneous linear and angular velocities of the sensor at time~\(t\). With our LOS definition~\(\boldsymbol{u}_i^t\) being the unit vector from the sensor origin to~\(\boldsymbol{p}_i^t\) (thus \(\boldsymbol{u}_i^t=\boldsymbol{p}_i^t/\|\boldsymbol{p}_i^t\|\)),
the radial projection of the rotational velocity about the sensor origin vanishes, i.e., \((\boldsymbol{u}_i^t)^\top(\boldsymbol{\omega}\times\boldsymbol{p}_i^t)=0\).
Hence, single-scan radial Doppler constrains \(\boldsymbol{v}\) but does not directly observe \(\boldsymbol{\omega}\). Rotation is estimated in Sec.~\ref{sec:match} through scan-to-scan registration using geometric constraints and the Doppler matching term.

We estimate~\(\boldsymbol{v}\) by robust regression over all points in the scan:

\begin{equation}
\min_{\boldsymbol{v}} 
\sum_{i} \rho\!\left( s_i^t + (\boldsymbol{u}_i^t)^\top \boldsymbol{v}  \right),
\label{eq:ego_lsq}
\end{equation}
with~\(\rho(\cdot)\) being a robust penalty (e.g., Huber). \eqref{eq:ego_lsq} is solved as a linear least squares in the three unknowns of~\(\boldsymbol{v}\)~\cite{wang2023infradar}.

Unlike previous ego motion estimators that employ global static assumptions~\cite{kellner2014instantaneous}~\cite{li20234d}, we suppose that a majority of points are static only at initialization. If \(t=0\), we set a neutral prior \(\boldsymbol{v}^{0}=\boldsymbol{0}.\)
For subsequent frames (\(t>0\)), we initialize the prediction from ego-velocity obtained from the previous registration
result~\(\hat{\boldsymbol{T}}_{t-1\rightarrow t}\). This policy reduces reliance on a persistent ``mostly static''
assumption beyond the first frame and accelerates convergence of \eqref{eq:ego_lsq}. At \(t=0\), Eq.~\eqref{eq:ego_lsq}
benefits from sufficient static support. If the sensor is initialized in a scene dominated by moving objects (e.g.,
heavy traffic with limited visible background), the initial estimate and thus the velocity filter can degrade for the
first frames. With the solution~\(\hat{\boldsymbol{v}}\), the velocity filter is defined as
\begin{equation}
r_i^t \;=\; s_i^t + (\boldsymbol{u}_i^t)^\top \hat{\boldsymbol{v}}.
\label{eq:residual}
\end{equation}
A point is flagged dynamic if~\(|r_i^t|>\tau(d_i)\), where~\(d_i=\|\boldsymbol{p}_i^t\|\) and~\(\tau(d)=\tau_0+\kappa d\) is a distance-adaptive threshold. The remaining points form the static set~\(\mathcal{S}_t\).
\subsection{Dynamic Point Clustering}\label{sec:cluster}
After computing the ego-velocity from static points, we use the dynamic points to cluster dynamic objects. Let~\(\mathcal{D}_t=\mathcal{P}_t\setminus\mathcal{S}_t\) be the set of dynamic candidates identified by the velocity filter. We employ spatial and velocity consistency analysis to cluster dynamic points.

Spatial consistency: We adopt HDBSCAN~\cite{campello2013density}, a hierarchical density-based clustering method robust to uneven point density and partial occlusion. We apply HDBSCAN to~\(\mathcal{D}_t\) with minimum cluster size~\(m_{\mathrm{cs}}\) (smallest object) and minimum samples~\(m_{\mathrm{s}}\) (controls density conservativeness and outlier robustness), producing clusters~\(\{\mathcal{C}_k^t\}\) and an explicit noise set~\(\mathcal{N}_t\).

\begin{figure}
    \centering
    \includegraphics[width=\linewidth]{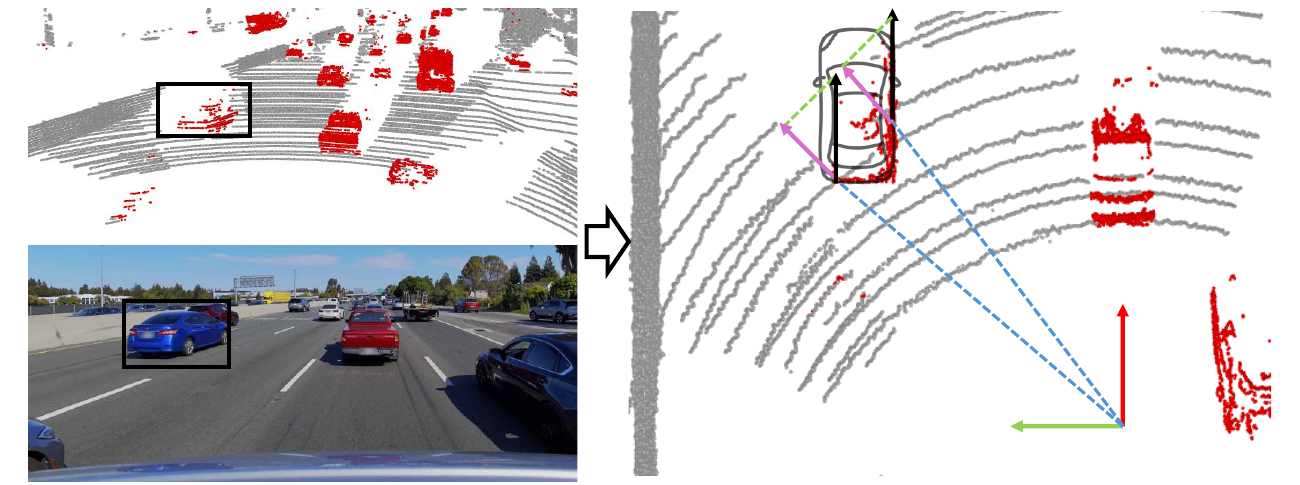}
    \caption{\small{Velocity reconstruction of dynamic objects. Left: FMCW LiDAR scan with dynamic points highlighted in \textcolor{red}{red} and synchronized camera view. Right: Top-down view of LiDAR scan. Dynamic points on the moving rigid body (car) satisfy the Doppler velocity consistency condition. This means that the velocity component of the object (black arrow) in the direction of the line-of-sight (\textcolor{blue}{blue} dashed line) equals the compensated Doppler velocity (\textcolor[HTML]{ff00ff}{purple} arrow) at that point.}}
    \label{fig:vel_con}
\end{figure}

Velocity consistency: For each cluster \(\mathcal{C}_k^t\), we assume that all points within the cluster belong to a rigidly moving object, as illustrated in Fig. ~\ref{fig:vel_con}. We first form the ego-motion–compensated Doppler velocity $\tilde{s}_i^t $ for each dynamic point~\(\tilde{s}_i^t\),
\begin{equation}
\tilde{s}_i^t \;\triangleq\; s_i^t + (\boldsymbol{u}_i^t)^\top \hat{\boldsymbol{v}},
\label{eq:vel_compensated}
\end{equation}
and recover the cluster velocity  $ \hat{\boldsymbol{v}}_k^t $ by ordinary least squares:
\begin{equation}
\varepsilon_i=(\boldsymbol{u}_i^t)^\top\hat{\boldsymbol{v}}_k^t-\tilde{s}_i^t,\qquad i\in\mathcal{C}_k^t,
\label{eq:vel_cluster}
\end{equation}
where~$ \varepsilon_i$ denotes the per-point velocity residuals and is then filtered by a velocity-adaptive threshold~$\tau_{v} = \lambda \cdot\hat{\boldsymbol{v}}$. We keep only the points satisfying
\begin{equation}
|\varepsilon_i|\;\le\;\tau_{v},
\label{eq:vel_gate}
\end{equation}
removing the rest to~\(\mathcal{N}_t\). If the inlier fraction falls below a threshold~\(\phi_{\min}\), the entire cluster is discarded. The 3D velocity reconstruction above requires sufficiently diverse LOS directions within a cluster. For distant or compact objects, the LOS vectors can become nearly parallel, making tangential components weakly observable and the least-squares problem ill-conditioned. We therefore assemble
\begin{equation}
\boldsymbol{U}_k^t \;\triangleq\;
\begin{bmatrix}
(\boldsymbol{u}_{i_1}^t)^\top\\
\vdots\\
(\boldsymbol{u}_{i_M}^t)^\top
\end{bmatrix}
\in\mathbb{R}^{M\times 3},
\qquad
\tilde{\boldsymbol{s}}_k^t \;\triangleq\;
\begin{bmatrix}
\tilde{s}_{i_1}^t\\
\vdots\\
\tilde{s}_{i_M}^t
\end{bmatrix},
\label{eq:Uk_def}
\end{equation}
where $\{i_m\}_{m=1}^M$ denotes the indices of the points in $\mathcal{C}_k^t$. We then estimate the cluster velocity by solving the linear system $\boldsymbol{U}_k^t \hat{\boldsymbol{v}}_k^t \approx \tilde{\boldsymbol{s}}_k^t$. To assess whether this system is numerically stable, we compute the singular values of $\boldsymbol{U}_k^t$. Let $\sigma_{\max}$ and $\sigma_{\min}$ be the largest and smallest singular values, and define the condition number $\kappa(\boldsymbol{U}_k^t)=\sigma_{\max}/\sigma_{\min}$. When $\sigma_{\min}$ is very small (nearly rank-deficient) or $\kappa(\boldsymbol{U}_k^t)$ is large (ill-conditioned), the tangential velocity components become unreliable. In these cases, we discard the cluster. This conservative filtering ensures that prediction is applied only to validated clusters, while the ego pose is estimated by scan-to-scan registration in Sec.~\ref{sec:match}.
After evaluating the spatial and velocity consistency of the clusters, the remaining points form refined clusters~\(\{\widetilde{\mathcal{C}}_k^t\}\). These clusters, along with their estimated velocities~\(\{\hat{\boldsymbol{v}}_k^t\}\), are used in subsequent prediction and matching stages.
\subsection{Dynamic Points Prediction}\label{sec:predict}
We predict only points belonging to dynamic clusters using a constant-velocity motion model. For each point~\(i\) in the refined dynamic clusters~\(\{\widetilde{\mathcal{C}}_k^t\}\) with translational velocities~\(\{\hat{\boldsymbol{v}}_k^t\}\),
\begin{equation}
\tilde{\boldsymbol{p}}_{i}^{\,t+1}
\;=\;
\boldsymbol{p}_i^t \;+\; \hat{\boldsymbol{v}}_k^t \Delta t\,,
\label{eq:pred_dynamic_only}
\end{equation}
which assumes cluster-wise constant translational velocity over~\(\Delta t\). The constant-velocity model sets the cluster’s angular rate to zero during prediction. If an object rotates with angular velocity~\(\boldsymbol{\omega}_k\), the unmodeled per-point displacement (relative to the cluster centroid~\(\boldsymbol{c}_k^t\)) is
\begin{equation}
\delta\boldsymbol{p}_{i}^{\mathrm{rot}}
\;\approx\;
\big(\boldsymbol{\omega}_k \times (\boldsymbol{p}_i^t-\boldsymbol{c}_k^t)\big)\,\Delta t .
\label{eq:rot_slippage}
\end{equation}
Because Doppler provides only radial velocity,~\(\boldsymbol{\omega}_k\) is not directly observable. We therefore (i) keep~\(\Delta t\) short, so~\(\|\delta\boldsymbol{p}_{i}^{\mathrm{rot}}\|\) remains small and unbiased at the centroid, (ii) use a slightly enlarged, distance-aware correspondence threshold for larger objects, and (iii) rely on the ICP point-to-plane term to absorb the residual rotational misalignment during pose refinement. In practice, this preserves accurate centroid prediction while bounding per-point errors by object size and rotation rate.
Points not in dynamic clusters are considered stationary during the sampling interval and thus are not predicted and remain at~\(\boldsymbol{p}_i^t\). The source set passed to matching is thus
\begin{equation}
\widetilde{\mathcal{P}}_{t}
=\Big\{\boldsymbol{p}_i^t \;\big|\; i\in\mathcal{S}_t\Big\}
\;\cup\;
\bigcup_k \Big\{\tilde{\boldsymbol{p}}_{j}^{\,t+1}\;\big|\; j\in\widetilde{\mathcal{C}}_k^{t+1}\Big\},
\label{eq:pred_set_dynamic_only}
\end{equation}
providing next-frame predictions for moving objects while leaving static structure unchanged for the subsequent Doppler-aware ICP alignment.
\subsection{Doppler-aware ICP Matching}\label{sec:match}
Given the predicted source set~\(\widetilde{\mathcal{P}}_{t}=\{(\boldsymbol{p}_i,\boldsymbol{u}_i,s_i)\}_{i=1}^{N_t}\) and the target set~\(\widetilde{\mathcal{P}}_{t+1}=\{(\boldsymbol{q}_j,\boldsymbol{n}_j,\boldsymbol{u}_j,s_j)\}_{j=1}^{N_{t+1}}\), we estimate the rigid motion~\(\boldsymbol{T}=[\boldsymbol{R}\,|\,\boldsymbol{t}]\in SE(3)\) with a two-term objective that combines geometry residual and Doppler residual. Here,~\(\boldsymbol{n}_j\in\mathbb{R}^3\) is the unit surface normal at~\(\boldsymbol{q}_j\) with it's neighbors, obtained by local plane fitting and normalized so that~\(\|\boldsymbol{n}_j\|=1\).

\begin{figure}
    \centering
    \includegraphics[width=\linewidth]{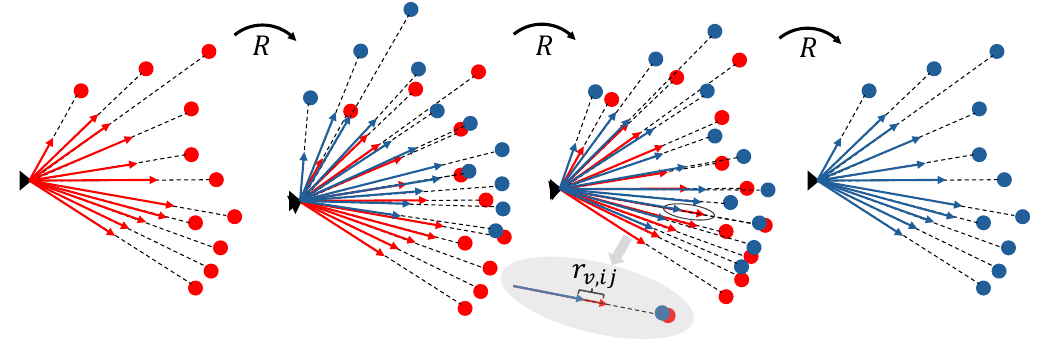}
    \caption{\small{Doppler residual of Doppler-aware ICP matching. \textcolor{red}{Red} points and arrows denote the source frame, and \textcolor{blue}{blue} points and arrows denote the target frame. Arrows show line-of-sight directions whose lengths are proportional to Doppler (radial) velocity. From left to right, the diagram illustrates the matching process: source Doppler rays are rotated to align with target rays, after which correspondences are established by combining point-to-plane geometry with the Doppler residual.}}
    \label{fig:dopplericp}
\end{figure}

Geometry residual: the point-to-plane term penalizes displacement along the target normal, encouraging the transformed source points to lie on the target surface and yielding fast, stable convergence in rigid registration~\cite{rusinkiewicz2001efficient}:
\begin{equation}
r_{g,ij} \;=\; \boldsymbol{n}_j^\top\big(\boldsymbol{R}\tilde{\boldsymbol{p}}_i+\boldsymbol{t}-\boldsymbol{q}_j\big).
\label{eq:ptp_final}
\end{equation}
Doppler residual: the Doppler term enforces agreement between the rotated source Doppler velocity and the target Doppler velocity along the line-of-sight, as illustrated in Fig.~\ref{fig:dopplericp}. It provides a complementary, rotation-focused cue that is independent of translation:
\begin{equation}
r_{v,ij} \;=\; \boldsymbol{u}_j^\top\,\boldsymbol{R}\big(s_i\,\boldsymbol{u}_i\big)\;-\; s_j.
\label{eq:doppler_final}
\end{equation}
We minimize a robust iteratively reweighted least squares objective
\begin{equation}
[\boldsymbol{R}\,|\,\boldsymbol{t}] = \min\sum_{(i,j)}
\Big[\;(1-\lambda_v)\rho_g\!\big(r_{g,ij}^2\big)\;+\;\lambda_v\,\rho_v\!\big(r_{v,ij}^2\big)\;\Big],
\label{eq:cost_final}
\end{equation}
where \(\rho_g,\rho_v\) are robust kernels (e.g., Huber) and \(0\leq\lambda_v\leq1\) balances the two terms. Since \(r_{v,ij}\) depends only on \(\boldsymbol{R}\) (its derivative with respect to \(\boldsymbol{t}\) is zero), it supplies a translation-invariant rotational constraint that (i) remains informative in repetitive or low-texture geometry (e.g., tunnels, bridges), (ii) is insensitive to depth ambiguity along the viewing ray (Fig.~\ref{fig:dopplericp}), and (iii) directly penalizes rotation errors when point-to-plane becomes weak or is affected by residual dynamics.
\section{Experimental Results}
\subsection{Implementation Details}
We implement Dynamic-ICP by extending the point-to-plane ICP pipeline from Open3D~\cite{Zhou2018} (KD-tree search, point-to-plane Jacobians) with an additional Doppler residual and its Jacobian. LOS directions~\(\boldsymbol{u}\) are stored separately from surface normals~\(\boldsymbol{n}\) to keep the geometric term strictly point-to-plane, while the Doppler term uses only LOS information. To limit the effect of outlier correspondences, we use Tukey loss on both terms, with tuning constants~\(\rho_g=0.5\) for the point-to-plane geometry residual (meters) and~\(\rho_v=0.3\) for the Doppler residual. We fix the balance to~\(\lambda_v=0.2\) across all experiments (chosen empirically). Dynamic clustering uses HDBSCAN with~\(m_{\mathrm{cs}}=30\) and~\(m_{\mathrm{s}}=10\), which control the smallest admissible object and density conservativeness, respectively. We keep these HDBSCAN parameters fixed across all datasets to avoid environment-specific tuning and to maintain a fully automatic and reproducible pipeline. Here, $m_{\mathrm{cs}}$ mainly controls the minimum object size (in number of points after voxelization), while $m_{\mathrm{s}}$ controls the conservativeness of clustering. We do not use additional sensors (IMU, GNSS, or wheel odometry). The initial pose is derived solely from the FMCW LiDAR: range supports geometric alignment, while per-point Doppler provides (i) an ego translational velocity estimate for dynamic filtering and (ii) an additional rotation-stabilizing cue through the Doppler matching term in Sec.~\ref{sec:match}.
\subsection{Experimental Evaluation}
We evaluate on four Doppler-capable datasets, focusing on highly dynamic segments:
\begin{itemize}
  \item HeLiPR~\cite{jung2024helipr}: driving sequences, recorded by an FMCW LiDAR.
  \item HeRCULES~\cite{hjkim2025icra}: diverse routes with fast ego movement and moving traffic, recorded by an FMCW LiDAR.
  \item AevaScenes~\cite{aevascenes}: highway and city scenes with two FMCW LiDARs.
  \item DICP~\cite{hexsel2022dicp}: curved walls (feature-less scenario) in CARLA Town05 with one simulated FMCW LiDAR.
\end{itemize}
We use the raw point clouds and raw per-point Doppler measurements as provided, without additional deskewing. For AevaScenes, which only supplies ego-motion compensated point clouds, we use these compensated point clouds directly. Moreover, since AevaScenes provides object-level velocity ground truth, we additionally validate the reconstructed object velocities against this reference. 
\subsection{Comparison to State-of-the-Art Methods}
\begin{figure*}[ht]
        \centering
        \includegraphics[width=1\linewidth]{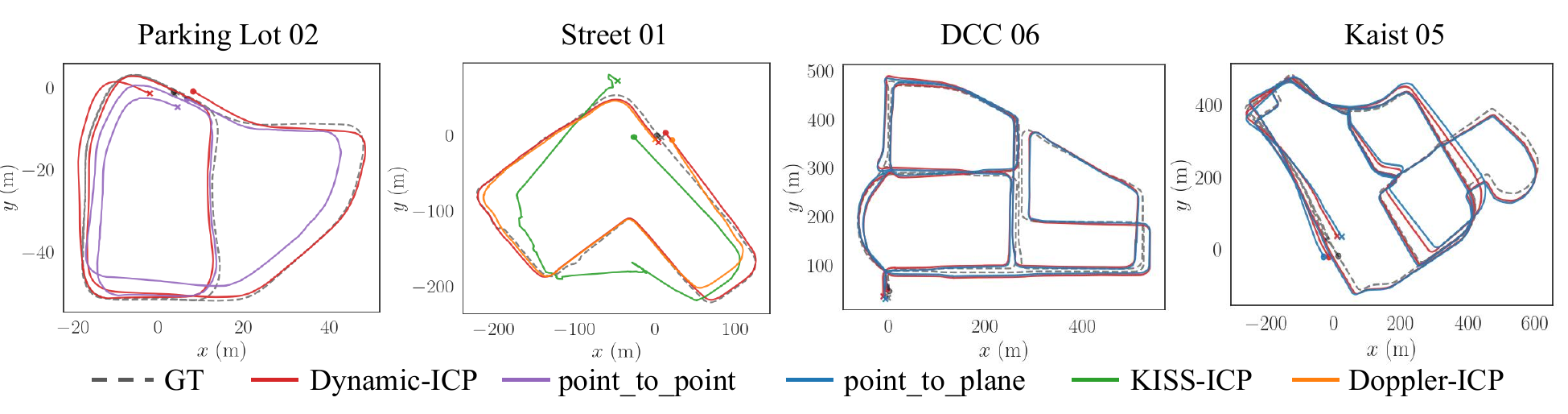}
        \caption{Qualitative comparison. For clarity, we plot only the two best-performing methods per sequence, including our Dynamic-ICP.}
        \label{fig:traj_plot}
    \end{figure*}

We benchmark against five state-of-the-art baselines under identical pre-processing (voxel grid, target normals) and correspondence radius. The baselines are: point-to-point ICP~\cite{besl1992method}, which minimizes Euclidean distances between corresponding points; point-to-plane ICP~\cite{rusinkiewicz2001efficient}, which projects errors onto target surface normals; Generalized-ICP~\cite{segal2009generalized}, a probabilistic point-to-plane formulation with local covariances; KISS-ICP~\cite{vizzo2023kiss}, a lightweight point-to-point ICP with adaptive correspondence thresholds; and Doppler-ICP~\cite{hexsel2022dicp}, which integrates Doppler velocity into point-to-plane ICP and requires sensor–vehicle extrinsic calibration. In contrast, our method does not require calibration between the sensor and the vehicle and is the only approach that performs cluster-wise dynamic prediction. Since we focus on registration accuracy, we report relative pose error (RPE) with respect to reference trajectories provided by the dataset at the frame gap, decomposed into rotation (RRE) and translation (RTE). The best and second best results are \textbf{bolded} and \underline{underlined}, respectively. Qualitative trajectories for the best-performing methods are shown in Fig.~\ref{fig:traj_plot}.

In the first experiment, we analyze the performance of our system and compare it to ICP-based methods on six HeRCULES sequences (Table~\ref{tab:hercules}). Dynamic-ICP delivers the best or tied-best performance across all scenes in both translation (RTE) and rotation (RRE) errors. It clearly improves rotation accuracy on challenging, repetitive geometry such as \textit{Bridge~01} and \textit{River Island 01}, while also reducing translation error (e.g., \textit{Library 01}). On \textit{Stream 01}, our method attains the lowest RTE and ties DICP for the best RRE. The only case without the top rotational score is \textit{Parking Lot 02}, where DICP and point-to-point outperform the other methods. The reason is that this is the only low ego-motion, near-static scenario, so our method degenerates into point-to-plane approach. This further demonstrates the advantage of our method in highly dynamic scenarios. Overall, Doppler-aware prediction and matching yield consistent orientation gains and competitive translation across diverse scenes.
\begin{table*}[ht]
    \setlength\tabcolsep{3pt}
    \centering
    \resizebox{\textwidth}{!}{
    \begin{tabular}{@{}l|cc|cc|cc|cc|cc|ccc@{}}\toprule
        \multirow{2}{*}{Method} & 
        \multicolumn{2}{c|}{\textbf{Bridge 01}} & 
        \multicolumn{2}{c|}{\textbf{Library 01}} & 
        \multicolumn{2}{c|}{\textbf{Parking Lot 02}} &
        \multicolumn{2}{c|}{\textbf{River Island 01}}  & 
        \multicolumn{2}{c|}{\textbf{Stream 01}} & 
        \multicolumn{2}{c}{\textbf{Street 01}}  
        \\ \cmidrule(l){2-13} 
        & RTE [m] &{RRE [\degrees]} & RTE [m] & {RRE [\degrees]} & RTE [m]  &{RRE [\degrees]} & RTE [m] &{RRE [\degrees]} & RTE [m] & {RRE [\degrees]} & RTE [m] & {RRE [\degrees]}      
        \\ \midrule
        ICP (point-to\_point) & 1.636 & 0.562 & 0.641 & 0.601 & 0.090 & \textbf{0.280} & 1.098 & 0.600 & 1.033 & 0.619 & 0.126 & \uline{0.150}
        \\
        ICP (point-to-plane) & 1.586	&  \uline{0.361} & 0.516 & 0.523 & \textbf{0.051} & 0.282 & 0.899 & 0.454 & 0.919 & 0.471 & 0.089 & 0.160
        \\ 
        GICP & 1.639 & 0.741 & 0.849 & 0.994 & 0.220 & 0.510 & 1.151 & 0.627 & 1.112 &	0.733 & 0.208 &	0.293
        \\ 
        KISS-ICP & 1.637 & 0.716 & 0.696 & 1.214 & 0.293 & 0.539 & 1.327 & 0.685 & 1.150 & 0.833 &	0.260 &	0.294
        \\ 
        Doppler-ICP & \uline{1.024} & 0.689 & \uline{0.178} & \uline{0.511} & 0.069 & \uline{0.281} & \uline{0.380} & \uline{0.334} & \uline{0.465} & \textbf{0.216} & \uline{0.071} &	\uline{0.150}
        \\  \midrule
        \textbf{Dynamic-ICP} & \textbf{1.023} & \textbf{0.248} & \textbf{0.177} & \textbf{0.285} & \textbf{0.051} & 0.282 & \textbf{0.379} & \textbf{0.189} &	\textbf{0.464} &	\textbf{0.216} &	\textbf{0.050} &	\textbf{0.140}
        \\ \bottomrule
    \end{tabular}}
    \caption{Comparison of ICP based methods on the HeRCULES dataset.}
    \label{tab:hercules}
\end{table*}
On HeLiPR dataset (Table~\ref{tab:helipr}), Dynamic-ICP achieves the best translation accuracy on five of six sequences and ties the remaining one, while consistently ranking first or second in rotation across all cases. It is particularly effective on scenes with weak geometric features or evolving appearance: \textit{Bridge 02} contains many moving vehicles with slow ego motion, \textit{Kaist 05} includes numerous pedestrians and long-term appearance changes, and \textit{Riverside 05} features many dynamic objects with long-term differences. Across these settings, motion prediction and Doppler-aware matching preserve reliable correspondences and stabilize orientation when static structure is limited or changing.
\begin{table*}[ht]
    \setlength\tabcolsep{3pt}
    \centering
    \resizebox{\textwidth}{!}{
    \begin{tabular}{@{}l|cc|cc|cc|cc|cc|ccc@{}}\toprule
        \multirow{2}{*}{Method} & 
        \multicolumn{2}{c|}{\textbf{Bridge 02}} & 
        \multicolumn{2}{c|}{\textbf{DCC 06}} & 
        \multicolumn{2}{c|}{\textbf{Kaist 05}} &
        \multicolumn{2}{c|}{\textbf{Riverside 05}}  & 
        \multicolumn{2}{c|}{\textbf{Roundabout 01}} & 
        \multicolumn{2}{c}{\textbf{Town 01}}  
        \\ \cmidrule(l){2-13} 
        & RTE [m] &{RRE [\degrees]} & RTE [m] & {RRE [\degrees]} & RTE [m]  &{RRE [\degrees]} & RTE [m] &{RRE [\degrees]} & RTE [m] & {RRE [\degrees]} & RTE [m] & {RRE [\degrees]}      
        \\ \midrule
        ICP (point-to\_point) & 0.671 & 0.252 & 0.381 & 0.302 & 0.324 & 0.278 & 0.923 & 0.366 & 0.386 & 0.249 & 0.370 & 0.167 \\
        ICP (point-to-plane) & \uline{0.534} & 0.195 & \uline{0.063} & \textbf{0.085} & \textbf{0.037} & \uline{0.102} & \uline{0.636} & \uline{0.232} & \textbf{0.105} & \uline{0.101} & 0.195 & \uline{0.105} \\
        GICP & 0.562 & 0.180 & 0.468 & 0.366 & 0.445 & 0.406 & 0.926 & 0.372 & 0.428 & 0.281 & 0.424 & 0.243 \\
        KISS-ICP & 0.558 & \textbf{0.100} & 0.237 & \uline{0.204} & \uline{0.215} & 0.266 & 1.143 & 0.682 & 0.222 & \textbf{0.095} & \uline{0.155} & 0.335 \\
        Doppler-ICP & 0.748 & 0.373 & 0.519 & 0.361 & 0.585 & 0.568 & 0.959 & 0.502 & 0.479 & 0.391 & 0.498 & 0.416 \\ \midrule
        \textbf{Dynamic-ICP} & \textbf{0.370} & \uline{0.110} & \textbf{0.062} & \textbf{0.085} & \textbf{0.037} & \textbf{0.101} & \textbf{0.592} & \textbf{0.150} & \uline{0.155} & 0.131 & \textbf{0.152} & \textbf{0.102} \\
        \bottomrule
    \end{tabular}}
    \caption{Comparison of ICP based methods on the HeLiPR dataset.}
    \label{tab:helipr}
\end{table*}
On the AevaScenes high-speed benchmark (Table~\ref{tab:icp_aeva}), we evaluate Dynamic-ICP on the full set of 44 highway sequences and 43 urban road sequences, and report the averaged frame-to-frame errors over all sequences. Overall, Dynamic-ICP remains accurate and stable across both highway and city driving, achieving the best performance in both translation and rotation on average. In addition to real-world data, we further conducted experiments on a simulation environment with degraded geometry (\textit{Curved Walls}), which mimics tunnel-like conditions with weak and repetitive structure. In this setting, Doppler-ICP achieves the best performance, which we attribute to its translation-coupled Doppler constraint that directly strengthens translation estimation under geometry degeneracy. Dynamic-ICP remains competitive and delivers consistently strong rotation accuracy, highlighting that the proposed Doppler-aware matching and motion prediction remain effective not only in highly dynamic traffic scenarios but also under challenging geometry. We additionally validate reconstructed cluster velocities on AevaScenes using object-level ground-truth annotations, obtaining low forward-axis error (MAE $0.53$\,m/s, median relative error $2.87\%$) and moderate lateral error (MAE $1.08$\,m/s), supporting reliable short-horizon prediction in real traffic. Table~\ref{tab:aeva_dynamic_stats} summarizes the degree of scene dynamics across representative sequences by reporting the ratio of dynamic candidates and the fraction of dynamic points that are clustered and retained for prediction. Overall, a substantial portion of the dynamic candidates can be consistently grouped and validated, indicating that the proposed dynamic clustering and refinement stage provides reliable support for motion prediction in real traffic scenes.
\begin{table}[htbp]
\centering
\resizebox{0.49\textwidth}{!}{%
\begin{tabular}{l|cc|cc|cc}
\toprule
\multirow{2}{*}{Method}  & \multicolumn{2}{c|}{\textbf{Highway 44}}& \multicolumn{2}{c|}{\textbf{City 43}} & \multicolumn{2}{c}{\textbf{Curved Walls}}\\ 
\cmidrule(l){2-7} & RTE [m] & RRE [°] & RTE [m] & RRE [°] & RTE [m] & RRE [°] \\
\midrule
ICP (point-to-point)  & \underline{0.974} & 0.168 & 0.703 & 0.160 & 0.555 & 0.226 \\
ICP (point-to-plane)  & 0.996 & 0.164 & 0.270 & \underline{0.067} & 0.263 & 0.067 \\
GICP                 & 1.052 & \underline{0.159} & 0.777 & 0.241 & 0.543 & 0.181 \\
KISS-ICP             & 0.985 & 0.256 & \underline{0.254} & 0.109 & 0.511 & 0.225 \\
Doppler-ICP          & 1.258 & 0.797 & 0.858 & 0.210 & \textbf{0.011} & \textbf{0.040} \\
\midrule
\textbf{Dynamic-ICP}  & \textbf{0.705} & \textbf{0.143} & \textbf{0.209} & \textbf{0.060} & \underline{0.063} & \underline{0.061} \\
\bottomrule
\end{tabular}%
}
\caption{Comparison of ICP based methods across sequences on AevaScenes dataset and DICP dataset.}
\label{tab:icp_aeva}
\end{table}
\begin{table}[htbp]
\centering
\resizebox{0.49\textwidth}{!}{%
\begin{tabular}{l|cc|c|c}
\toprule
\multirow{2}{*}{Sequence} &
\multicolumn{2}{c|}{Dyn./All} &
Clustered/Dyn. &
Refined/Dyn. \\
\cmidrule(l){2-3}
& Max & Mean & Mean & Mean \\
\midrule
highway\_day\_1    & 35.7\% & 22.3\% & 71.1\% & 95.8\%  \\
city\_day\_1   & 33.9\% & 17.6\% & 70.3\% & 93.4\%  \\
Bridge 02    & 29.2\% & 3.9\% & 71.0\% & 89.6\% \\
Street 01    & 16.8\%  & 7.6\% & 77.4\% & 93.1\% \\
\bottomrule
\end{tabular}%
}
\caption{\small{Dynamic-scene statistics on HeRCULES, HeLiPR, and AevaScenes. Dyn./All: dynamic candidates over all scan points; Clustered/Dyn.: clustered dynamics over dynamic candidates; Refined/Dyn.: validated dynamics used for prediction. Max/Mean are computed over frames in each sequence.}}
\label{tab:aeva_dynamic_stats}
\end{table}
\subsection{Ablation Studies}
We conduct additional experiments to investigate the individual benefits of three core components of Dynamic-ICP: velocity filter (VF), dynamic points prediction (DPP), and the Doppler residual (DR) by deactivating each while holding all other settings fixed. As shown in Table~\ref{tab:ablation_core}, disabling the velocity filter means we no longer separate dynamic from static points and, consequently, do not run dynamic points prediction. Dynamic points remain in the “static” set and corrupt correspondences, increasing both RTE and RRE. Removing dynamic points prediction means we still detect dynamics with the velocity filter, but discard them and register using only static points. Dropping Doppler residual reduces the objective to standard point-to-plane ICP, removing the rotation-only cue. Orientation becomes less stable in repetitive or low-texture scenes, increasing RRE even when geometry is well aligned. The full model (VF + DPP + DR) consistently yields the lowest errors.
\begin{table}[htbp]
\centering
\resizebox{0.49\textwidth}{!}{%
\begin{tabular}{l|cc|cc|cc}
\toprule
\multirow{2}{*}{Method}  & \multicolumn{2}{c|}{\textbf{River Island 01}}& \multicolumn{2}{c|}{\textbf{Riverside 05}} & \multicolumn{2}{c}{\textbf{Highway 44}}\\ 
\cmidrule(l){2-7}
        & RTE [m] & {RRE [\degrees]} & RTE [m] & {RRE [\degrees]} & RTE [m] & {RRE [\degrees]} \\
        \midrule
        w/o VF  & 0.871 & 0.680 & 0.686 & 0.235 & 1.078 & 0.268 \\
        w/o DPP & \uline{0.411} & \uline{0.346} & 0.633 & 0.237 & \uline{0.899} & 0.172 \\
        w/o DR  & 0.900 & 0.552 & \uline{0.595} & \uline{0.154} & 0.971 & \uline{0.161} \\ \midrule
        \textbf{Full} & \textbf{0.379} & \textbf{0.189} & \textbf{0.592} & \textbf{0.150} & \textbf{0.705} & \textbf{0.143} \\
        \bottomrule
\end{tabular}%
}
\caption{Ablation evaluation on Dynamic-ICP.}
\label{tab:ablation_core}
\end{table}

\begin{table}[ht]
    \setlength\tabcolsep{6pt}
    \centering
    \begin{tabular}{@{}lccc@{}}
        \toprule
        Method & Avg. iters $\downarrow$ & Conv. rate $\uparrow$ & FPS $\uparrow$ \\
        \midrule
        ICP (point-to-point)   & 22.94 & 98.00\%           & 13.02 \\
        ICP (point-to-plane)   & 17.20 & \uline{99.70\%}   & \textbf{15.10} \\
        GICP                     & 23.65 & 98.90\%           & 5.68 \\
        Doppler-ICP              & \uline{13.54} & \textbf{99.80\%} & \uline{14.20} \\
        \textbf{Dynamic-ICP}     & \textbf{13.50} & \textbf{99.80\%} & 13.34 \\
        \bottomrule
    \end{tabular}
    \caption{Runtime and convergence statistics on HeRCULES.}
    \label{tab:runtime_conv}
\end{table}
Table~\ref{tab:runtime_conv} compares efficiency and robustness across methods. The \emph{convergence rate} is defined as the percentage of registrations that terminate within $I_{\max}=100$ iterations and satisfy the threshold $\epsilon=10^{-5}$ (with a maximum point correspondence distance of $0.3$\,m for all methods). Overall, Dynamic-ICP converges in fewer iterations than classic ICP variants and similarly to Doppler-ICP, while achieving the highest throughput and matching the best convergence rate. These results indicate that Doppler cues and dynamic point prediction stabilize optimization, reduce iterations, and enable real-time performance without sacrificing reliability.
\section{CONCLUSION}
We presented Dynamic-ICP, a training-free, Doppler-aware registration framework that employs scene motion as an additional signal for scan registration. Our approach (i) estimates ego translational velocity from per-point Doppler and builds a velocity filter, (ii) clusters moving points with HDBSCAN and reconstructs object-wise 3D velocities, (iii) predicts dynamic points with a constant-velocity model, and (iv) aligns scans using a compact objective that combines point-to-plane geometric residual with a translation-invariant, rotation-only Doppler residual. Unlike Doppler-ICP variants that require sensor–vehicle extrinsic calibration, our approach operates without external sensors or calibration, relying solely on FMCW LiDAR range and Doppler.


\bibliographystyle{plain_abbrv}
\bibliography{bibliography,this}
\end{document}